\documentclass[a4paper, 10pt, conference]{ieeeconf}      
\usepackage{FG2025}

\FGfinalcopy 

\usepackage{url}            
\usepackage{booktabs}       
\usepackage{amsfonts}       
\usepackage{nicefrac}       
\usepackage{microtype}      
\PassOptionsToPackage{table}{xcolor}
\usepackage{soul}
\usepackage{wrapfig}
\usepackage{hyperref}

\usepackage{amsmath}
\usepackage{adjustbox}
\usepackage{makecell}
\usepackage{multirow}
\usepackage{subcaption}

\newcommand{\ie}{\textit{i.e.}}
\newcommand{\eg}{\textit{e.g.}}
\newcommand{\wrt}{\textit{w.r.t.} }

\definecolor{lightyellow}{rgb}{0.95,0.95,0.8}
\sethlcolor{lightyellow}

\makeatletter
\newcommand{\thickhline}{%
    \noalign {\ifnum 0=`}\fi \hrule height 1.1pt
    \futurelet \reserved@a \@xhline
}

\def\Cline#1#2{\@Cline#1#2\@nil}
\def\@Cline#1-#2#3\@nil{%
  \omit
  \@multicnt#1%
  \advance\@multispan\m@ne
  \ifnum\@multicnt=\@ne\@firstofone{&\omit}\fi
  \@multicnt#2%
  \advance\@multicnt-#1%
  \advance\@multispan\@ne
  \leaders\hrule\@height#3\hfill
  \cr}
\makeatother

\newcolumntype{"}{@{\hskip\tabcolsep\vrule width 1.1pt\hskip\tabcolsep}}
\makeatother

\IEEEoverridecommandlockouts                              
\def\FGPaperID{80} 

\title{\LARGE \bf
 3D Face Reconstruction Error Decomposed: \\A Modular Benchmark for Fair and Fast Method Evaluation
 }

\author{\parbox{16cm}{\centering
    {\large Evangelos Sariyanidi$^1$, Claudio Ferrari$^{2,3}$, Federico Nocentini$^4$, Stefano Berretti$^4$,\\ Andrea Cavallaro$^{5,6}$ and Birkan Tunc$^{1,7}$}\\
    {\normalsize
    $^1$ The Children’s Hospital of Philadelphia, USA\\
    $^2$ Department of Architecture and Engineering, University of Parma, Italy\\
    $^3$ Department of Information Engineering and Matemathics, University of Siena, Italy \\
    $^4$ Media Integration and Communication Center (MICC), University of Florence, Italy\\
    $^5$ Idiap Research Institute, Switzerland\\
    $^6$ École Polytechnique Fédérale de Lausanne (EPFL), Switzerland\\
    $^7$ University of Pennsylvania, USA\\}}
}

\usepackage{fancyhdr}
\begin{document}

\ifFGfinal
\thispagestyle{empty}
\pagestyle{empty}
\else
\author{Anonymous FG2025 submission\\ Paper ID \FGPaperID \\}
\pagestyle{plain}
\fi
\maketitle


\begin{abstract}
Computing the standard benchmark metric for 3D face reconstruction, namely geometric error, requires a number of steps, such as mesh cropping, rigid alignment, or point correspondence. Current benchmark tools are monolithic (they implement a specific combination of these steps), even though there is no consensus on the best way to measure error. We present a toolkit for a Modularized 3D Face reconstruction Benchmark (M3DFB), where the fundamental components of error computation are segregated and interchangeable, allowing one to quantify the effect of each. Furthermore, we propose a new component, namely {\em correction}, and present a computationally efficient approach that penalizes for mesh topology inconsistency. Using this toolkit, we test 16 error estimators with 10 reconstruction methods on two real and two synthetic datasets. Critically, the widely used ICP-based estimator provides the worst benchmarking performance, as it significantly alters the true ranking of the top-5 reconstruction methods. Notably, the correlation of ICP with the true error can be as low as 0.41. Moreover, non-rigid alignment leads to significant improvement (correlation larger than 0.90), highlighting the importance of annotating 3D landmarks on datasets. Finally, the proposed correction scheme, together with non-rigid warping, leads to an accuracy on a par with the best non-rigid ICP-based estimators, but runs an order of magnitude faster. Our open-source codebase is designed for researchers to easily  compare alternatives for each component, thus helping accelerating progress in benchmarking for 3D face reconstruction and, furthermore, supporting the improvement of learned reconstruction methods, which depend on accurate error estimation for effective training. 
\end{abstract}

\section{Introduction}\label{sec:intro}
The standard benchmark metric for 3D reconstruction is the geometric error between the reconstructed mesh and the corresponding ground truth. However, the computation of this error is not trivial, as the compared meshes underlie different topologies. Therefore, one needs to perform a number of steps prior to computing error, such as rigid or non-rigid alignment of the meshes, or establishing point correspondence. As a result, all computed errors are mere estimators of the true error, and their quality depends on all the involved steps. 

Standard benchmark procedures are typically built on a monolithic pipeline that depends on a specific choice of components, even when there is no research to back them up. Moreover, recent studies suggest that commonly used alignment procedures such as the Iterative Closest Point (ICP) algorithm~\cite{chen1992object} or point correspondence approaches~\cite{amberg2007optimal,myronenko2010point} may be inadequate~\cite{ferrari2021sparse}, to the point of altering the true ranking of compared methods~\cite{sariyanidi23}. Thus, there is a need to determine the best benchmarking practices in the current state of the art and improve on them.

\begin{figure*}[t!]
    \centering
    \includegraphics[width=0.95\linewidth]{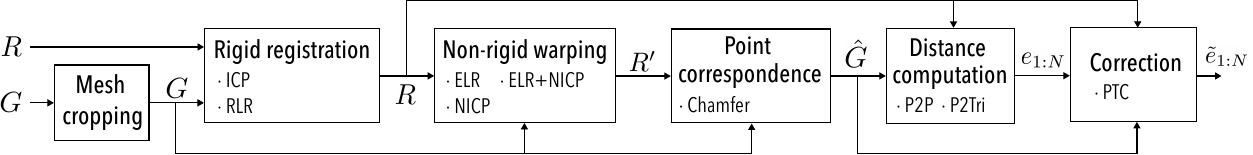}
    \caption{The main steps involved in the computation of the geometric error between reconstructed meshes ($R$) and ground truth scans ($G$). Note that mesh cropping, non-rigid warping, and correction are optional steps. The methods that our toolkit implements for each step are listed next to a bullet point ($\cdot$)}
    \label{fig:main}
\end{figure*}

This paper presents a Modularized 3D Face reconstruction Benchmark (M3DFB)\----a toolkit where the fundamental components are segregated and easily interchangeable (Figure~\ref{fig:main}). The proposed toolkit enables one to measure the effect of each step involved in error computation, and identify the procedures and components that yield the best error estimates. We also introduce an additional step, namely {\em correction}, as we show that the error estimate can be improved without computing new correspondences. The toolkit is structured in a way that proposing alternatives for each component is simple. Furthermore, the code aims at efficiency with caching and parallelization; and contains built-in tools for visualization and scientific reporting (Appendix~A). 

Using our framework, we test 16 estimators with 10 reconstruction methods on two real and two synthetic datasets, where the latter are used to measure the real error and evaluate the reliability of estimators~\cite{sariyanidi23}. Results lead to new insights for fair comparison of reconstruction methods and suggestions for better dataset curation; and challenge some practices of conventional wisdom~(Section~\ref{sec:discussion}).
In summary, the main contributions of this paper are as follows:
\begin{itemize}
    \item A modular toolkit to implement existing procedures and to easily propose new alternatives.
    \item A new step, namely {\em correction}, and a computationally efficient scheme that leads to error estimation on a par with non-rigid ICP (NICP).
    \item Using the modularity of the framework, we compare a large number of benchmark procedures, providing novel insights that highlight good practices and expose limitations.
\end{itemize}
The open-source codebase of M3DFB is made available at \url{https://github.com/sariyanidi/M3DFB}, with the goal of facilitating continued progress in benchmarking metrics and procedures for 3D face reconstruction. 

\section{Related Work and Motivation}
\label{sec:related}
The most widely used approach for measuring reconstruction error is computing the average point-to-point nearest-neighbor distance after rigidly registering reconstructed and ground truth meshes via ICP~\cite{jackson17,deng2019accurate,gecer2019ganfit,gecer2021fast,genova18,Bai_2021_CVPR, prnet, feng2018evaluation, sariyanidi20, shang20}, landmarks~\cite{liu2018disentangling,zhu2020beyond} or a combination of both. Even official benchmarks such as NoW~\cite{RingNet:CVPR:2019} or~\cite{feng2018evaluation} use this type of approach. Zhu \textit{et al.}~\cite{yang2020facescape} use a different strategy for their FaceScape dataset, performing rigid alignment after projecting the two meshes on the image plane via a fixed perspective camera model. The recently proposed REALY benchmark~\cite{chai2022realy} uses a more sophisticated solution based on NICP~\cite{amberg2007optimal}, to establish region-based point correspondences and compute local errors. Here, a question arises: is there a way to say which one among all the possible options is more reliable? A recent paper~\cite{sariyanidi23} proposed to answer this question via experiments on synthetic data, and showed that, while NICP leads to more accurate error estimation with respect to a simple rigid alignment, there is still need for improvement. Moreover, NICP adds a significant computational overhead, as it can take minutes per mesh to compute~\cite{chai2022realy,sariyanidi23}, which limits applicability, particularly in the case of internal validation (\eg, parameter tuning) that requires repeated evaluation. Also, it must be noted that the afore-mentioned study~\cite{sariyanidi23} did not investigate all components of the error computation pipeline. For example, rigid alignment was done via landmarks, even though ICP is also widely used. 

Our modularized toolkit is motivated by the fact that there is need for more accurate and computationally efficient estimators of geometric error; and that the estimated error is expected to be sensitive to all components in Figure~\ref{fig:main}. A toolkit that allows the systematic comparison of all possible error estimators facilitates the discovery of the ones that are most promising and accurate. Furthermore, an expandable toolkit makes it easier for researchers to implement novel error estimators and compare them with alternatives. Further, we add a novel step to the error computation pipeline, namely correction. This step aims to improve estimated error without computing new correspondences, and we show that this is possible by penalizing for mesh topology inconsistency~(Section~\ref{sec:pipeline}).

Using the proposed toolkit, we perform, to our knowledge, the largest systematic comparison of error estimators. Our experiments expose two critical findings. First, the widely popular ICP-based approach mentioned above emerges as the worst error estimator. Second, it is possible to estimate error with an accuracy on a par with the best NICP-based estimators yet without the computational overhead of the latter. Experiments on two distinct types of mesh topologies, namely that of the Basel Face Model (BFM)~\cite{bfm09} and FLAME~\cite{flame}, highlight the generalizability of the tested estimators. Our procedures do not require dataset-specific annotations (\eg, dense patches~\cite{chai2022realy}), but only a sparse set of landmarks on the face~(Figure~\ref{fig:chamfer_vs_ELR}b), and thus applicable to a wide set of datasets, such as Florence~\cite{bagdanov2011florence}, BU4DFE~\cite{yin20063d}, Bosphorus~\cite{savran2008bosphorus}, FRGC~\cite{phillips2005overview} or the newly released Florence expression dataset~\cite{ferrari2023florence}.

\section{The Proposed Framework}
\label{sec:pipeline}
Our toolkit computes geometric error between a reconstructed mesh $R\in \mathbb{R}^{N\times 3}$, and the corresponding ground truth $G \in \mathbb{R}^{M\times 3}$ through six steps, three of which are optional (Figure~\ref{fig:main}). Users can implement any combination of steps simply by curating a JSON. More details on the implementation and structure of the framework are provided in Appendix~A. The six steps are summarized in the following sections. 

\subsection{Mesh cropping} 
A ground truth $G$ obtained with real sensors can contain points from irrelevant parts (\eg, hair, shoulders) or clothes. The face or head region is typically cropped out of $G$ by eliminating points farther than a pre-determined distance from a reference point, \eg, the nose tip. 

\subsection{Rigid registration} 
In general, $R$ and  $G$ are in different coordinate systems. Thus, a rigid alignment is needed and is typically performed through a similarity transformation (\ie, scale, rotation and translation) on the reconstructed mesh. Some recent studies also aimed at measuring the metric (\ie, absolute) error by omitting scale transformation~\cite{zielonka22}. 

The transformation that is needed for rigid registration can be computed in two ways. First, by using the ICP algorithm, which performs dense registration by using all points in $R$ and $G$, or a uniformly sampled subset of these points. The second alternative is rigid landmark-based registration (RLR), which uses a sparse subset of available landmarks on the ground truth. At minimum, three points are needed for the landmark-based alignment, although more points can be included for a more robust registration. Both ICP and RLR options are included in our toolkit. 


\subsection{Non-rigid warping}
\begin{figure*}[t]
\centering
\includegraphics[width=0.80\linewidth]{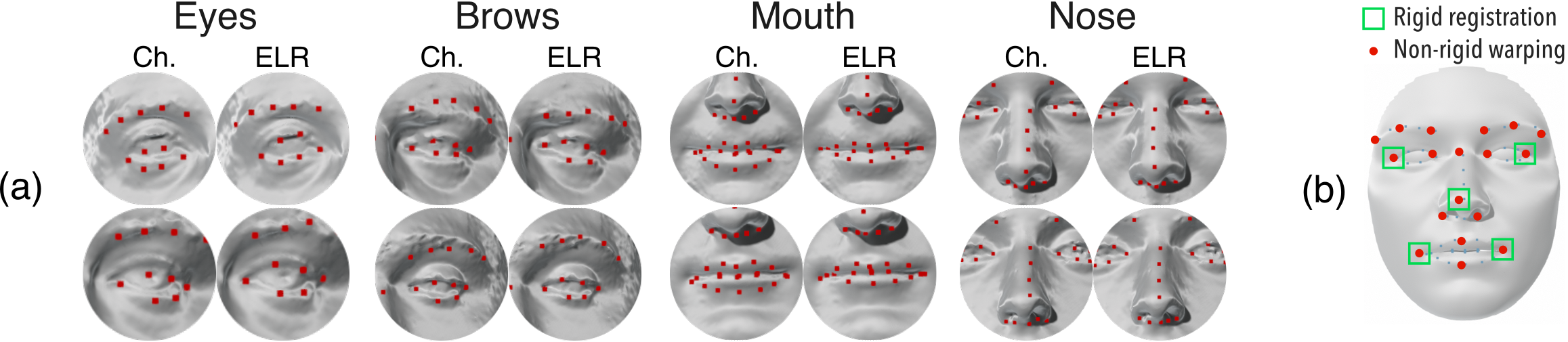}
\caption{(a) Incorrect Chamfer~(\ie, nearest-neighbor) correspondences (Ch.) and the improvement achieved by non-rigid warping via ELR, illustrated by overlaying (red) landmark points from the reconstructed mesh $R$ that correspond to eyes, brows, mouth and nose on corresponding ground truth scans. (b) Landmark points used as reference for rigid- and non-rigid alignment.}
\label{fig:chamfer_vs_ELR}
\end{figure*}

Depending on the quality of the reconstruction $R$, the rigidly aligned meshes can have a significant amount of mismatch.  For example, Figure~\ref{fig:chamfer_vs_ELR}a shows that the points of $R$ that are supposed to enclose the eye region of $G$ can fail to do so. Mismatches of this kind can be problematic for point correspondence the following point~\emph{d}.

In such cases, one can reduce the mismatch by applying non-rigid alignment~(\eg, see ELR in Figure~\ref{fig:chamfer_vs_ELR}a). The non-rigidly warped version of $R$, denoted as $R'$, is used only for establishing point correspondence, and not while computing the per-vertex error~(Figure~\ref{fig:main}). 

One way to perform non-rigid warping is to use an NICP algorithm, which can optionally be guided by some landmark points. NICP is thus far the most used approach for non-rigid alignment~\cite{booth20163d,chai2022realy,ploumpis2020towards}. However, NICP leads to a significant computational overhead~(Section~\ref{sec:related}). Another drawback is that NICP has several hyper-parameters to tune, and the optimal configuration might depend on the mesh topology or ground truth density. Ultimately, results computed by different research groups might be severely altered if different setups or NICP variants are used. Finally, it was shown that NICP can fail in matching semantically relevant points~\cite{ferrari2021sparse}. 

Another alternative for non-rigid warping is to use Elastic Landmark-based Registration (ELR; Appendix~B), referred to also as landmark-based pre-alignment~\cite{sariyanidi23}. This process is computationally efficient, as it is based on solving a closed-form linear system. However, ELR requires landmark annotations on the ground truth that may not be available on a dataset. Annotating these in a fully or semi-automated way is not difficult when the ground truth scans include texture information, as the latter can be used to render multiple images and leverage efficient 2D landmark detection algorithms~\cite{bulat2017far}. Both NICP and ELR options, as well as their combination (ELR+NICP), are included in our toolkit.

\subsection{Point correspondence} 
A ground truth obtained with a real 3D scanner has a topology (\ie, number and density of points) that is not known a priori and differs from that of the reconstructed mesh. As such, one must establish point correspondence between $R$ and $G$ prior to error computation. That is, for each point (\ie, row) $r_i$ of $R$, one must find the index $i'$ of the corresponding point (\ie, row) $g_{i'}$ of $G$. As a result of this process, we obtain the \textit{matched ground truth} $\hat{G} \in \mathbb{R}^{N\times 3}$. That is, the $i$th row of $\hat{G}$, namely $\hat{g}_i$, is the point that is estimated to correspond to the point $r_i$ of $R$.

The standard criterion for point correspondence is the Chamfer criterion, which is based on finding the ground truth point that is closest (\ie, nearest neighbor) to each point on the reconstructed mesh. Another variant of this approach is to find the closest point along a ray defined by the surface normals, but it was shown to only provide slight improvement for high curvature regions~\cite{pan2013establishing}. Also, while there exist learned point correspondence approaches~(\eg,~\cite{urbach2020dpdist}), these are proposed for estimating shape rather than for benchmarking purposes. In sum, we implement only the Chamfer approach.

\subsection{Per-vertex distance computation} 
\label{sec:distance_comp}
Once point correspondence is established, the error for each vertex of $R$ can be computed as the point-to-point (P2P) distance $e_i:=|| r_i - \hat{g}_i||.$ One may also compute alternative distances, such as the point-to-plane distance as in~\cite{liu2020learning,luo2021normalized}. Our toolkit implements the latter as a point-to-triangle (P2Tri) distance, which is computed by identifying the triangle on $\hat G$ corresponding to each point in $r_i$. 

\begin{figure}[t!]
    \centering
    \includegraphics[width=0.65\linewidth]{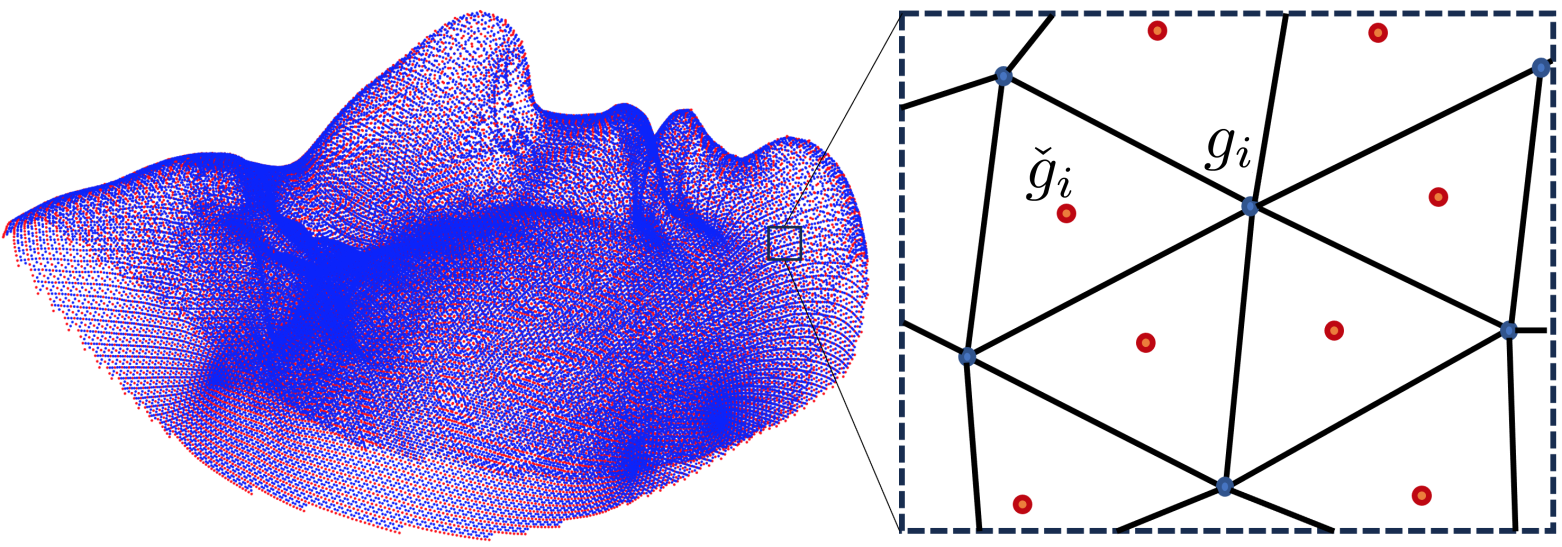}
    \caption{Re-meshing: blue points belong to $G$, re-meshed (red) points $\check{g}_i$ are obtained as barycenters of facets of $G$}
    \label{fig:remeshing}
\end{figure}

We define this triangle with the three vertices on $\hat G$ that are closest to $r_i$. We do not use surface normals, as they can be sensitive to noise in the ground truth scan. 

The P2Tri distance can be useful when the compared reconstruction methods rely on inconsistent mesh topologies, as meshes with different topologies can have a non-negligible P2P error even if they represent practically the identical surface. As an example, consider the two overlaid meshes in Figure~\ref{fig:remeshing}. Both represent the same surface, but through a different triangularization. Specifically, the vertices of the red mesh are the barycenters of the facets of the original (\ie, blue) mesh points. On an experiment that we conducted on 100 synthesized meshes, the average P2P error between such re-parametrized meshes was 0.71mm, which is larger than the error difference between many actual reconstruction methods (Section~\ref{sec:results}), whereas the P2Tri distance correctly captured that the meshes represent the same surface with an error that was practically zero~($6.9e^{-5}$). However, further preliminary experiments highlighted that the P2Tri distance does not improve benchmarking performance on meshes parametrized in the same way~(Table~\ref{tab:pointvsplane}). Given that our experiments are designed to compare meshes with the same topology, we only report the P2P distance as in most literature works, although both P2P and P2Tri are included in our toolkit.

\begin{table}[!t]
\centering
\caption{Error comparison between point-to-point (P2P) and point-to-triangle (P2Tri) metrics both when the correspondence is known (True error) and estimated.}
\small
\begin{tabular}{c|c||c|c|c|c|c}
\hline
Rec. method & & P2P & P2Tri & & P2P & P2Tri\\ 
\hline \hline
3DI-m & \multirow{4}{*}{\rotatebox[origin=c]{90}{True Error}} & \textbf{2.08}	&  \textbf{1.86} & \multirow{4}{*}{\rotatebox[origin=c]{90}{Chamfer}} & \textit{1.39} & \textit{1.26} \\
Deep3D-m &	&	\textit{2.10}	& \textit{1.85} &  & \textbf{1.33} & \textbf{1.20} \\
3DDFAv2-m &	&	\underline{2.47} & \underline{2.18} & &	1.51 &  1.39\\
INORig-m &	&	2.66	& 2.35 & &	\underline{1.47} & \underline{1.34} \\ 
\hline 
\hline
\end{tabular}
\label{tab:pointvsplane}
\end{table}


\subsection{Correction} 
Our toolkit contains a novel step towards the estimation of geometric error, namely correction. The purpose of this step is to compute an updated per-vertex error $\tilde{e}_i$ without computing new correspondences. We propose a simple and generically applicable correction scheme that improves estimated error with little computational overhead.

\begin{figure}[t!]
\centering
\includegraphics[width=0.9\linewidth]{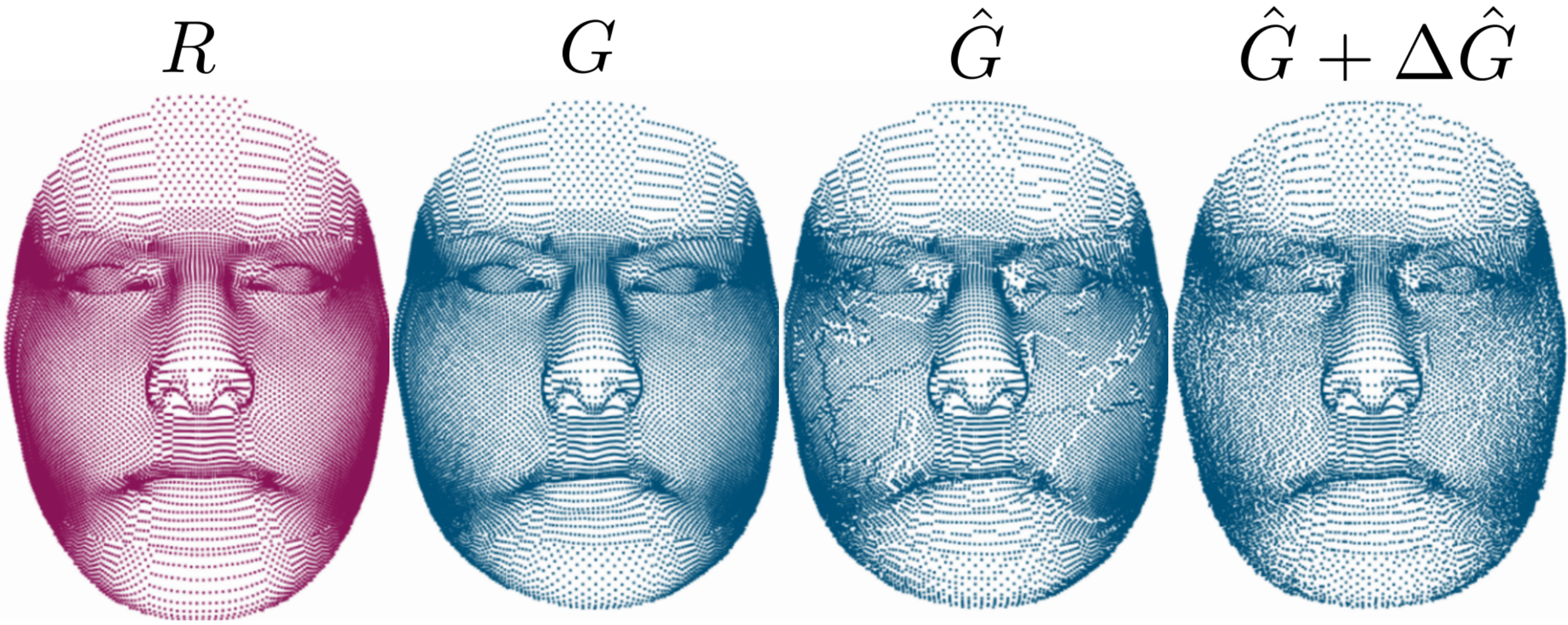}
\caption{Illustration of ETC. Incorrect Chamfer-based point correspondences between $R$ and $G$ often lead to visible gaps and fractures in the matched mesh $\hat G$. The correction term $\Delta \hat G$ exploits this by penalizing for such gaps.}
\label{fig:correction}
\end{figure}

The proposed correction scheme is motivated by the following observation. The point correspondence established with nearest neighbor or similar criteria typically leads to a matched ground truth mesh, $\hat G$, that has gaps and fractures that are not seen on $R$ or $G$~(Figure~\ref{fig:correction}). Such gaps and fractures occur because two or more distinct points on $R$ are incorrectly assigned to the same point or to very close points on $G$. Experiments on synthetic data show that 10\% (Appendix~C) or more mesh points have incorrect duplicate correspondences, and that this can depend on the method, compromising the fairness of an experimental comparison.

We propose a correction scheme to improve the estimate of geometric error by \text{enforcing (mesh) topology consistency} (ETC), as shown in Figure~\ref{fig:correction}. Suppose that $g'_i$ is the correct point corresponding to $r_i$. Our goal is to compute a correction term $\delta\hat g_i$ for each matched ground truth point $\hat g_i$ such that the updated average error, namely
\begin{equation}
\frac{1}{N}\sum_{i=1}^N \tilde e^{ETC}_i:=\frac{1}{N}\sum_{i=1}^N||r_i-(\hat g_i-\delta\hat g_i)|| ,
\label{eq:correction}
\end{equation}
is a better predictor of the actual error $\frac{1}{N}\sum_i ||r_i-g'_i||$ compared to $\frac{1}{N}\sum_{i}||r_i-\hat g_i||$. We hypothesize that this goal will be achieved if the correction terms make the spacing among the corrected ground truth points $\hat g_i+\delta\hat g_i$ to be similar to the spacing among the corresponding points on $R$, reducing the visible gaps and fractures~(Figure~\ref{fig:correction}). This can be achieved by solving a quadratic optimization problem, which can be done efficiently by solving a sparse linear system. Below, we describe how such a correction term be computed. For simplicity, we compute the correction terms separately for the $x$, $y$ and $z$ axes of the mesh points. 

Let $e_i$ be the $i$th unit vector in $\mathbb R^3$, and $P_x$ be the $N\times N$ permutation matrix such that the entries of the vector $\tilde r_x = P_x R e_1$ are sorted in ascending order. Further, let $\tilde g_x$ be the vector of the matched ground truth points permuted in the same way, $\tilde g_x=P_x \hat G e_1$. Our goal is to find a vector $\delta \tilde g_x \in \mathbb R^N$ such that the spacing between neighboring points of the reconstructed mesh (\ie, the difference between consecutive entries of $\tilde r_x$), is similar to the difference between the consecutive points on $\tilde g_x+\delta\tilde g_x$. This can be achieved by minimizing the function:
\begin{equation}
\frac{1}{2}\delta\tilde g^T_x D^T D \delta \tilde g_x - \epsilon_x^T D^T D \delta \tilde g_x ,
\end{equation}

\noindent
\wrt $\delta \tilde g_x$, where $\epsilon_x=\tilde r_x-\tilde g_x$, and $D$ is the $(N-1)\times N$ matrix, whose $i$th column has non-zero values only at the $i$th and $(i+1)$th entries: 
\begin{equation}
    D = \begin{pmatrix}
        1 & -1 & 0 & \dots   & 0  & 0\\
        0 &  1 & -1 & \dots  & 0 & 0\\
        \vdots & \vdots & \vdots & \ddots & \vdots & \vdots \\
        0 &  0 & 0 & \dots  & 1 & -1\\
    \end{pmatrix}.
\end{equation}

Without additional constraints or terms, this is an ill-posed problem, as $D^T D$ is rank deficient---its rank is $N-1$. To have a well-defined problem, we apply $\ell_2$ regularization on $\delta \tilde g_x$, which amounts to restricting the movement of each point by a weight $w_i$. The weight of a point $w_i$ can be determined based on how confident we are about the correspondence established for the point. If the correspondence is correct, then the correction term should be small. Arguably, the only criterion that we can use to attribute confidence to a point correspondence is the proximity of the point to the $L$ annotated landmarks, particularly if a non-rigid warping such as ELR was applied. Thus, we determine each weight $w_i$ by its proximity to landmarks. Specifically, the proximity of the ground truth point $\hat g_i$ to the landmarks is measured simply as the average of the distance to the closest landmark $h_1(\hat g_i)=\min_j||\hat g_i-g_{l_j}||$, and the average distance to landmarks $h_2(\hat g_i)=\sum_j ||\hat g_i-g_{l_j}||/L$ after removing the minimum of the latter and normalizing \wrt the inter-ocular distance $d_{\text{iod}}$ as:
\begin{equation}
    w_i = (h_1(\hat g_i) + h_2(\hat g_i)-\min_j h_2(\hat g_j))/2 d_{\text{iod}} .
\end{equation}

\noindent
When we use the $\ell_2$ regularization described above, the correction term $\delta \tilde g^*_x$ is the minimizer of the function:
\begin{equation}
f(\delta \tilde g_x; \epsilon_x) = \frac{1}{2}\delta\tilde g^T_x Q \delta \tilde g_x - \epsilon_x^T D^T D \delta \tilde g_x,
\end{equation}

\noindent
where $Q = D^T D+W$, and $W$ is a diagonal matrix whose diagonal entries are $w^2_1, w^2_2, \dots, w^2_N$. This is a convex optimization problem as $Q$ is positive definite. Furthermore, the problem can be solved efficiently even for large $N$, because the Cholesky factors of $Q$ can be computed straightforwardly due to the sparsity of $D$. The same procedure is used for the $y$ and $z$ axes, to define permutation matrices $P_y$ and $P_z$, and estimate terms $\delta\tilde g_y^*$ and $\delta\tilde g_z^*$. Finally, the correction term $\Delta \hat G$ is computed as:
\begin{equation}
\Delta \hat G = \begin{pmatrix}
P^T_x\delta\tilde g_x^* & P^T_y\delta\tilde g_y^* & P^T_z\delta\tilde g_z^* 
\end{pmatrix}, \label{eq:correction_term}
\end{equation}

\noindent
where the transposed permutation matrices $P_x^T$, $P_y^T$, and $P_z^T$ are applied to recover the ordering of row indices that is consistent with $\hat G$ and $R$. The estimated error for a given mesh $R$ is then computed through~\eqref{eq:correction}, where $\delta \hat g_i$ is simply the $i$th row of $\Delta\hat G$.

\section{Criteria for evaluating a benchmark}\label{sec:criteria-benchmark}
Accurate and fair benchmarks are paramount for continued progress in research. Tasks such as classification, retrieval or object detection come with well-defined ground-truths and metrics (\eg, classification accuracy, Receiver Operating Characteristic curves, Sensitivity or Recall) that have a unique definition and can be computed directly from predictions and ground truths; thus, they do not need complicated benchmarks, and researchers can even copy-paste results directly from previous papers without worrying about discrepancies. 
In 3D face reconstruction, this does not apply, as there are many ways to compute the error, raising the question of how to choose the benchmark metric. When designing our benchmark framework, we identified three crucial characteristics that a reliable metric should adhere to.

\noindent
\textbf{Estimation accuracy:} This is the most important aspect of a benchmark. Given that the computed error is inevitably an estimate, the setting that is most faithful to the real error should be prioritized. Currently, the only way to get this information is by using synthetically generated meshes with a fixed and known topology. This eliminates the need for all the pre-processing steps, including the point correspondence estimation. This can be regarded as the \textit{true} error and we use it as a reference when evaluating an error estimator.

\noindent
\textbf{Computational efficiency and consistency:} even though this might seem a minor issue, computing the error should not take a reasonable amount of time or resources. Running NICP, for example, takes minutes per mesh, hence estimating the error on hundreds of samples might become unbearable. Clearly, accuracy should be the main focus, yet efficient solutions would be preferable. Moreover, the number of tunable parameters should be minimized, to maximize consistency across different implementations or runs. NICP, for example, has several hyper-parameters that needs to be tuned, which can change results significantly.

\noindent
\textbf{Dataset generalization:} A benchmark metric should be general, and not tailored to a specific test dataset. In some cases, \eg, challenges, one can define its own framework, yet in the most general case benchmarks should be applicable to any dataset. The recent REALY~\cite{chai2022realy} benchmark for example is limited in this sense as it requires per-region annotations, preventing its use with other datasets.

\section{Experiments}
\label{sec:experiments}
We conduct experiments with 16 error estimators to identify (combinations of) error computation steps that lead to increased accuracy. We experiment on real data to inspect the behavior of error estimators on data obtained with real scanners and on synthesized data to compare estimated and true error~\cite{sariyanidi23}. To observe whether compared estimators are sensitive to the topology of reconstructed meshes, we use reconstruction methods based on two morphable models, \ie, BFM and FLAME. 

\subsection{Experimental setup}
\label{sec:experimental_setup}
\paragraph{Tested error estimators}
We test 16 error estimators by considering all possible combinations of the two rigid registration (ICP, RLR), four non-rigid warping (none, ELR, NICP, ELR+NICP), and two correction options (none, ETC). These estimators are referred to as E1\--16 (Table~\ref{tab:main}). All estimators use the P2P distance, as our experiments are based on comparing reconstruction methods with the same mesh topology~(Section~\ref{sec:pipeline}). RLR is applied with five landmark points~(Figure~\ref{fig:chamfer_vs_ELR}b). When ICP is used for rigid registration, we still apply RLR for initialization, in which case we observe that the need for mesh cropping is obviated in our experiments. The framework is implemented in python. ICP is implemented using the \texttt{open3d} library, and the NICP implementation is taken from the REALY benchmark~\cite{chai2022realy}.

\paragraph{Datasets}
To compare estimated errors with the true error, we generate synthetic data~\cite{sariyanidi23}. We evaluate BFM-based reconstruction methods with data generated from the BFM, and FLAME-based methods with data generated from FLAME. Both synthesized datasets include 100 synthesized identities, and we render 50 (neutral) face images per identity from different poses. We also report the performance of error estimators on two real datasets, namely Florence~\cite{bagdanov2011florence} and BU4DFE~\cite{yin20063d}, by similarly rendering 50 frames per subject from (neutral) scans. 
The landmarks needed for rigid registration or non-rigid warping (Figure~\ref{fig:chamfer_vs_ELR}b) on real datasets are computed automatically with the approach used by Booth~\textit{et al.}~\cite{booth20163d}. That is, we render multiple 2D images for each scan mesh; then detect 2D landmarks~\cite{bulat2017far} per frame and refine them by fitting a morphable model~\cite{sariyanidi24}; and finally re-map the landmarks to 3D and average across images.

\begin{table*}
    \caption{Comparison of the true geometric error of several reconstruction methods (left column) with error estimated via 16 estimators (E1\---E16). Estimators differ in their rigid alignment (ICP or RLR) or Non-rigid Warping (ELR, NICP, or ELR+NICP) approach; or in the usage of correction. Methods ending with "-m" indicate multi-frame variants~(Section~\ref{sec:experimental_setup}). Highlighted cells indicate that the ranking of a reconstruction method is estimated incorrectly. ${}^*$Estimator uses ETC correction.}
\small
\def\arraystretch{1.2}
\setlength{\tabcolsep}{0.20em} 
    \centering
    \begin{tabular}{llccc|cc|cc|cc"cc|cc|cc|cc} \Xcline{4-19}{1.1pt}
 &  &\parbox[t]{2mm}{\multirow{4}{*}{\rotatebox[origin=c]{90}{True Error}}}
 & \multicolumn{8}{c"}{ICP} & \multicolumn{8}{c}{RLR}  \\ \cline{4-19}
  & & & \multicolumn{2}{c|}{\----} & \multicolumn{2}{c|}{ELR} & \multicolumn{2}{c|}{NICP} & \multicolumn{2}{c"}{\makecell{ELR+\\NICP}} & \multicolumn{2}{c|}{\----} & \multicolumn{2}{c|}{ELR} & \multicolumn{2}{c|}{NICP} & \multicolumn{2}{c}{\makecell{ELR+\\NICP}} \\\cline{4-19}
& & &E1&E2${}^*$&E3 &E4${}^*$&E5&E6${}^*$&E7&E8${}^*$&E9&E10${}^*$&E11&E12${}^*$&E13&E14${}^*$&E15&E16${}^*$\\ \thickhline
\parbox[t]{2.4mm}{\multirow{8}{*}{\rotatebox[origin=c]{90}{BFM methods}}} &3DI-m       &2.34&\hl{1.21}&\hl{1.76}&1.89&2.29&\hl{1.76}&\hl{1.82}&2.01&2.06&1.43&2.03&1.84&2.27&2.02&2.07&2.09&2.14 \\
&Deep3D-m    &2.53&\hl{1.19}&\hl{1.65}&2.25&2.51&\hl{1.58}&\hl{1.64}&2.16&2.20&1.61&2.15&2.24&2.53&2.12&2.17&2.28&2.32\\
&Deep3D      &2.69&\hl{1.27}&1.76&2.43&2.70&\hl{1.78}&\hl{1.84}&2.34&2.38&1.69&2.26&2.39&2.70&2.26&2.31&2.44&2.48\\
&INORig-m    &3.04&\hl{1.19}&\hl{1.68}&2.46&2.72&\hl{1.70}&\hl{1.77}&2.36&2.41&1.72&2.34&2.43&2.76&2.27&2.32&2.49&2.53\\
&3DDFA-m     &3.06&1.33&1.88&2.72&3.00&1.92&1.97&2.63&2.67&1.92&2.58&2.72&3.03&2.61&2.65&2.79&2.83\\
&3DDFA       &3.09&\hl{1.33}&\hl{1.88}&2.75&3.02&1.94&2.00&2.66&2.69&\hl{1.92}&\hl{2.57}&2.74&3.04&2.62&2.66&2.81&2.85\\
&Synergy     &3.65&1.46&2.08&3.23&3.50&2.29&2.34&3.14&3.17&2.11&2.82&3.20&3.51&2.99&3.02&3.28&3.31\\
&3DI         &4.10&1.93&2.60&3.49&3.81&3.12&3.16&3.59&3.62&2.36&3.10&3.48&3.84&3.53&3.57&3.71&3.75\\
\Xhline{1.1pt}
\parbox[t]{2.4mm}{\multirow{6}{*}{\rotatebox[origin=c]{90}{FLAME methods}}} &MICA-m      &2.43&1.32&1.80&2.37&2.63&1.93&2.05&2.22&2.33&1.49&2.08&2.41&2.71&2.22&2.32&2.46&2.56 \\
&MICA        &2.62&1.42&1.94&2.55&2.82&2.14&2.25&2.44&2.55&1.59&2.22&2.57&2.88&2.41&2.52&2.66&2.76 \\
&DECA        &2.83&1.43&1.95&2.66&2.89&2.32&2.42&2.65&2.75&1.68&2.26&2.83&3.08&2.62&2.71&2.93&3.02 \\
&RingNet     &3.01&1.48&2.01&2.90&3.13&2.37&2.48&2.81&2.91&1.88&2.50&3.04&3.31&2.74&2.83&3.09&3.17 \\
&RingNet-m   &3.02&\hl{1.48}&\hl{2.01}&\hl{2.90}&\hl{3.13}&2.39&2.50&2.82&2.92&\hl{1.88}&\hl{2.50}&3.05&\hl{3.31}&\hl{2.74}&\hl{2.83}&3.10&3.18 \\
&EMOCA       &3.17&\hl{1.45}&\hl{2.00}&3.05&3.28&NA&NA&NA&NA&\hl{1.74}&\hl{2.33}&3.24&3.48&NA&NA&NA&NA \\ \thickhline
    \end{tabular}
    \label{tab:main}
\end{table*}

\paragraph{Implemented reconstruction methods}
We implement the BFM-based reconstruction methods 3DDFAv2~\cite{guo2020towards}, 3DI~\cite{sariyanidi24}, Deep3D~\cite{deng2019accurate}, INORig~\cite{Bai_2021_CVPR} and SynergyNet~\cite{wu21}; and the FLAME-based methods RingNet~\cite{RingNet:CVPR:2019}, DECA~\cite{feng21}, MICA~\cite{zielonka22} and EMOCA~\cite{danecek22}. For methods that perform multi-frame reconstruction or can be extended to multi-frame naïvely~\cite{genova18}, we also report multi-frame reconstruction results (from 50 frames). 

\paragraph{Evaluation metrics}
We measure the accuracy of an estimator by computing the average estimated per-subject geometric error on synthetic data, and comparing it with the true average per-subject error. For a fair comparison, per-subject error is computed as the average per-vertex error on mesh points that are common to the compared methods. BFM-based methods produce reconstructions cropped in different ways, and we use the 23,470 mesh points common to all methods. Additionally, we report error for the 14,000 points close to facial landmarks (Figure~\ref{fig:correlations}). For FLAME-based methods, we use the indices corresponding to face points, discarding other regions (\eg, scalp, neck, or ears). We further validate an error estimator by computing the rate of inconsistency~\cite{sariyanidi23}, which measures the fairness of an estimator by quantifying the degree to which the estimator over- or under-estimates error inconsistently across reconstruction methods. Finally, we compute the correlation between the true and estimated errors of methods.

\subsection{Results}
\label{sec:results}
\paragraph{Comparison with true error on synthetic data}
Table~\ref{tab:main} compares the output of 16 error estimators with the true error on synthetic data~\cite{sariyanidi23}, separately for BFM- and FLAME-based reconstruction methods. The widely popular error estimator based on ICP and Chamfer correspondence~(Section~\ref{sec:related}), namely E1, provides the worst performance. On BFM data, E1 incorrectly ranks the top five reconstruction methods, and adding ETC correction (\ie, E2) does not improve this outcome. Moreover, using NICP does not guarantee a substantial improvement, as the estimators E5 and E6 incorrectly rank the first four BFM-based methods. However, the performance of NICP can be significantly improved with the initialization: when rigid alignment is performed with RLR, or when the (non-rigid) ELR warping is included, the ranking consistency of NICP improves notably (see E7, E8, E13-E16). 

\begin{figure*}
    \centering
    \includegraphics[width=0.8\linewidth]{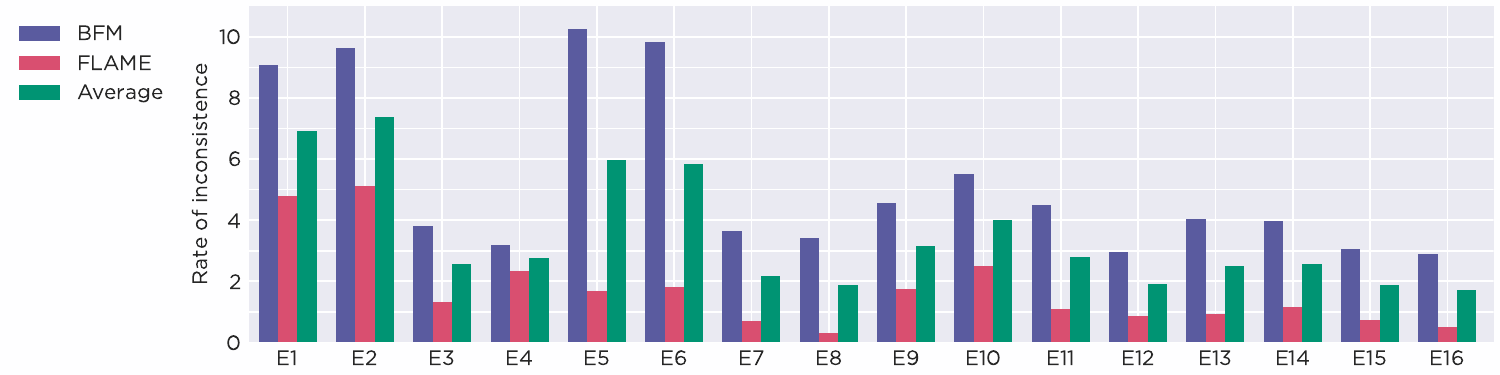}
    \caption{Rate of inconsistency of compared estimators on BFM and FLAME datasets, and the average of the two rates (the lower the better).}
    \label{fig:ROIs}
\end{figure*}

Overall, the most consequential component in Table~\ref{tab:main} is ELR. Whenever the latter is used, all BFM-based methods are ranked correctly, and FLAME-based methods are ranked with no more than one mistake between two very similarly performing methods (RigNet and RingNet-m). While there exist a number of estimators that correctly rank (nearly) all reconstruction methods, it must be noted that the performance gap is not always estimated accurately. For example, the true error difference between Deep3D and INORig-m is 0.35mm, yet it is estimated to be as small as 0.01mm (E14) or, at most, 0.08mm (E10). Thus, to reveal more nuanced differences between error estimators, we invoke the rate of inconsistency metric (the smaller the better; see Section~\ref{sec:experimental_setup}).

Figure~\ref{fig:ROIs} shows the rate of inconsistency of all compared error estimators on BFM and FLAME data, as well as the average on the two. On average, the five best performing estimators are E7, E8, E12, E15 and E16. Four of these estimators use ELR+NICP, and one (E12) uses ELR alone. Of note, E12 relies on correction, and removing it~(\ie, E11) leads to a nearly 50\% increased rate of inconsistency. That E12 is comparable to the other best estimators is noteworthy, as the non-rigid warping step that it underlies is an order of magnitude faster than NICP~(see computational efficiency results in Table~\ref{tab:comp_time}).

\begin{table}[!h]
    \centering
    \caption{Computation time (seconds) of error estimation components.}
    \begin{tabular}{cccccc} \thickhline
        RLR&ICP&ELR &NICP&Chamfer&ETC\\ \hline
        0.01&0.30&3.46&53.28&14.84&0.12\\ \thickhline
    \end{tabular}
    \label{tab:comp_time}
\end{table}

\begin{figure*}
    \centering
    \includegraphics[width=0.95\linewidth]{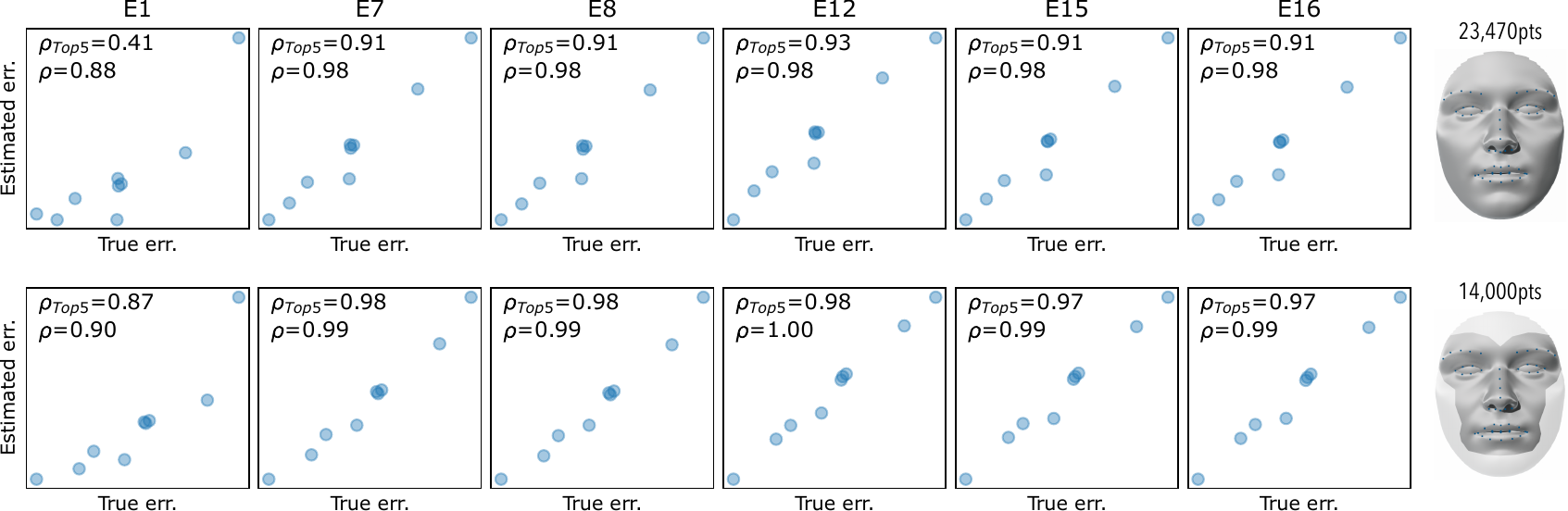}
    \caption{True vs. estimated error for the estimators E1, E7, E8, E12, E15 and E16 shown for all eight BFM-based methods in Table~\ref{tab:main} as well as mean-face estimation (\ie, the error between the mean face of BFM and the ground truths). Correlation is computed from all reconstruction methods ($\rho$) or from the five best methods $(\rho_{Top5})$. Top row: error  estimated from all 23,470 mesh points; bottom row: error estimated from the 14,000 points close to landmarks.}
    \label{fig:correlations}
\end{figure*}

\begin{figure}[!h]
\centering
\includegraphics[width=0.7\linewidth]{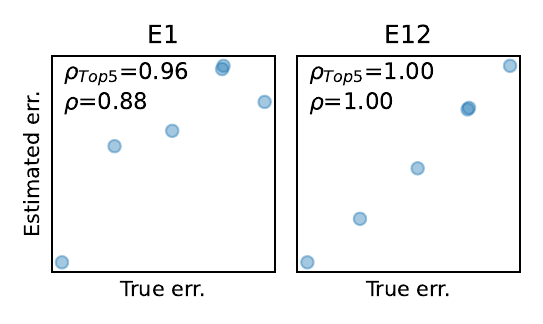}
\caption{True vs. estimated error for E1 and E12 for FLAME methods.}
\label{fig:correlations_FLAME}
\end{figure}

The performance of the afore-mentioned error estimators is illustrated further in Figure~\ref{fig:correlations} (top) by plotting the true vs. estimated errors for BFM-based reconstrution methods, and reporting their correlation. The correlation for E1 is only 0.41 when we focus on the top-5 reconstruction methods, whereas all other estimators achieve a correlation of at least 0.91. Since the ELR step of the latter estimators is based on landmarks, one may expect that per-vertex error is estimated with higher accuracy for mesh vertices close to the landmarks. Indeed, the correlation of all estimators other than E1 is at least 0.97 when error is computed from the 14k points corresponding to inner face~(Figure~\ref{fig:correlations}, bottom). The performance of E12 on FLAME-based reconstruction methods~(Figure~\ref{fig:correlations_FLAME}) indicates that it works similarly on meshes with a very different mesh topology as well, suggesting that the underlying ELR and ETC correction steps are robust to mesh topology variations.

\begin{figure}
    \centering
    \includegraphics[width=0.99\linewidth]{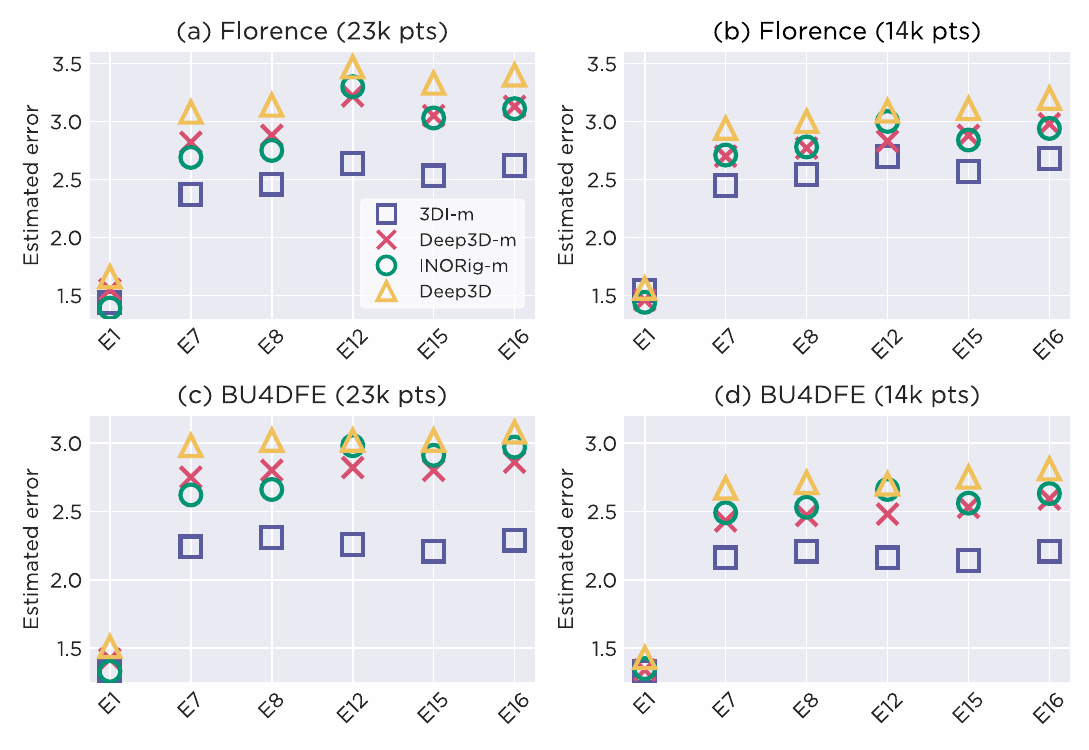}
    \caption{Estimated error on the Florence and BU4DFE data. Results are reported for all 23k mesh points that are contained commonly in all compared methods, as well as a subset of 14k points that correspond to the inner face.}
    \label{fig:real_errors}
\end{figure}

\paragraph{Comparison on real data}
Figure~\ref{fig:real_errors} shows the estimated reconstruction error on two real datasets, namely Florence and BU4DFE. Error is reported with the five best estimators from the previous section (E7, E8, E12, E15, E16) as well as the standard E1 approach.  For clarity, results are shown only for the four best reconstruction methods.

According to the E1 estimator, all reconstruction methods perform very similarly across Figure~\ref{fig:real_errors}(a--d). However, all other estimators show visible differences between methods, which, in light of the previous section, probably means that E1 artificially squashes the performance of compared methods and alters their ranking. All other estimators are consistent in their prediction of the reconstruction method that produces the highest and smallest error, although the differences between the 2nd and 3rd best methods is difficult to surmise, particularly when the entire meshes with 23k points are used. The performance of all reconstruction methods on Florence and BU4DFE are reported in Table~\ref{tab:florence_results} and~\ref{tab:BU4DFE_results}.

\begin{table}[h!]
    \centering
    \caption{Estimated geometric error (mm) of reconstruction methods on Florence data, computed with six estimators. Methods above (below) the double-line are based on the BFM (FLAME) mesh topology. The first, second, and third best methods are respectively denoted with bold, italic, and underlined text. NA indicates that error failde to be computed due to NICP.}
    \begin{tabular}{l|c|c|c|c|c|c} \thickhline
    & E1 & E7 & E8 & E12 & E15 & E16 \\ \hline
3DI-m       &\textit{1.44}&\textbf{2.37}&\textbf{2.46}&\textbf{2.64}&\textbf{2.53}&\textbf{2.62}\\
Deep3D-m &\underline{1.55}&\underline{2.82}&\underline{2.88}&\textit{3.22}&\underline{3.05}&\underline{3.13}\\
Deep3D &1.66&3.08&3.14&3.47&3.33&3.40\\
INORig-m    &\textbf{1.39}&\textit{2.69}&\textit{2.75}&\underline{3.30}&\textit{3.03}&\textit{3.11}\\
3DDFAv2-m   &1.56&3.03&3.08&3.53&3.29&3.34\\
3DDFAv2   &1.58&3.05&3.11&3.56&3.31&3.37\\
SynergyNet&1.70&3.31&3.36&3.72&3.52&3.57\\
3DI      &1.69&3.19&3.24&3.57&3.36&3.42 \\ \hline\hline
MICA-m   &\textbf{1.40}&\textbf{3.59}&\textbf{3.75}&\textbf{3.99}&\textbf{4.14}&\textbf{4.28} \\
MICA   &\textit{1.46}&\textit{3.70}&\textit{3.85}&\textit{4.08}&\textit{4.26}&\textit{4.39} \\
DECA   &\underline{1.62}&\underline{3.91}&\underline{4.07}&\underline{4.24}&\underline{4.35}&\underline{4.49} \\
RingNet-m &1.86&4.52&4.68&4.79&4.72&4.87 \\
RingNet &1.86&4.54&4.69&4.79&4.73&4.87 \\
EMOCA  &1.73&NA&NA&4.57&NA&NA \\ \thickhline
    \end{tabular}
    \label{tab:florence_results}
\end{table}

\begin{table}[h!]
    \centering
    \caption{Estimated geometric error (mm) of compared reconstruction methods on the BU4DFE dataset, computed with six different estimators analyzed in Section~4. Methods above (below) the double-line are based on the BFM (FLAME) mesh topology. The first, second, and third best methods are respectively denoted with bold, italic, and underlined text. NA indicates that error computation failed due to the NICP component.}
    \begin{tabular}{l|c|c|c|c|c|c} \thickhline
    & E1 & E7 & E8 & E12 & E15 & E16 \\ \hline
3DI-m       &\textit{1.54}&\textbf{2.45}&\textbf{2.54}&\textbf{2.70}&\textbf{2.57}&\textbf{2.68} \\
Deep3D-m &\underline{1.46}&\textit{2.70}&\textit{2.77}&\textit{2.83}&\underline{2.88}&\underline{2.98} \\
Deep3D &1.56&2.94&3.00&3.09&3.11&3.20 \\
INORig-m    &\textbf{1.44}&\underline{2.71}&\underline{2.78}&\underline{3.00}&\textit{2.84}&\textit{2.94} \\
3DDFAv2-m   &1.67&3.17&3.24&3.48&3.25&3.33 \\
3DDFAv2    &1.70&3.22&3.28&3.54&3.30&3.39 \\
SynergyNet &1.84&3.75&3.81&3.83&3.72&3.8 \\
3DI      &1.75&3.27&3.33&3.59&3.39&3.48 \\ \hline\hline
MICA-m   &\textit{1.98}&\textbf{4.61}&\textbf{4.75}&\textbf{4.42}&\textbf{4.97}&\textbf{5.09}\\
MICA   &\underline{2.03}&\underline{4.70}&\underline{4.84}&\underline{4.52}&\textit{5.05}&\textit{5.17}\\
DECA   &\textbf{1.95}&\textit{4.66}&\textit{4.79}&\textit{4.37}&5.15&\underline{5.26}\\
RingNet-m &2.10&4.99&5.13&4.65&\underline{5.14}&5.27\\
RingNet &2.10&5.00&5.14&4.66&5.15&5.28\\
EMOCA  &2.04& NA&NA&4.72&NA&NA \\ \thickhline
    \end{tabular}
    \label{tab:BU4DFE_results}
\end{table}

\section{Limitations}
The computation of true error can only be done when the reconstructed and ground truth meshes share the same topology. Thus, our comparisons were intra-topology; BFM-based and FLAME-based methods were ranked separately, and neither our results nor any other comparisons in the literature to our knowledge can currently ensure the fairness of cross-topology experiments. While the P2Tri error may eliminate small differences in mesh topologies, such as uniform mesh re-parametrization or downsampling (Section~\ref{sec:distance_comp}), it cannot be claimed that it can eliminate more fundamental differences in mesh topologies such as those BFM and FLAME (\eg, uniform vs. non-uniform mesh density). 

The ETC component that we introduced typically improves results only if non-rigid warping is performed. This implies that a good initialization to point correspondence is necessary for ETC to work efficiently. Nevertheless, ETC is merely one way of implementing the correction step that we introduced in this paper, and future research can yield alternative methods for correction that are more robust to initialization.

\section{Discussion and Conclusions}
\label{sec:discussion}
We proposed a modularized framework for estimating 3D face reconstruction error by decoupling the fundamental steps involved in its computation. This modular design allowed us to unveil critical limitations of current benchmarks, and analyze the impact of each step involved in the process. Some critical findings that emerged are listed below.

The widespread practice of estimating error with ICP and then Chamfer correspondences is inadequate, as it works particularly poorly with the best reconstruction methods. Furthermore, while using NICP improves performance,  results are sensitive to initialization. Of note, our analysis exposed an alternative to NICP-based error estimation that works significantly faster, namely ELR followed by correction. To the best of our knowledge, this is the first validated and computationally efficient approach to geometric error estimation that works comparably to NICP. Another critical finding is that using landmark-based non-rigid warping significantly and consistently improves performance. Thus, we advise the collection of data with texture scans, as annotating landmarks on shape alone is ambiguous and  requires laborious manual intervention. 

Overall, a number of estimators outperformed significantly the standard approach (ICP+Chamfer), yet there is no conclusive evidence to suggest which one is the best. Arguably, a reasonable strategy is to use multiple reliable estimators until a consensus can be arrived at. We believe that our open-source M3DFB framework can enable further progress by making it easier to consistently compare results from different research groups, expose the limitations of existing benchmark metrics, and facilitating the implementation and experimental validation of new error estimation procedures. Besides, other than for benchmarking purposes, getting the error right is of utmost importance for learned reconstruction methods as well, which need correct error computation for effective training. Our framework can serve to the latter aim as well, by providing a tool for designing effective loss functions for training learned methods.  

\section*{Ethical Statement}
In this paper, we proposed a modularized benchmarking procedure for accurately evaluating and fairly comparing 3D face reconstruction methods. The outcomes and results of our work are unlikely to lead to negative societal effects, or harm to anyone. This is because, despite involving human data, \ie, face images and 3D scans, all the data used in our research comes from publicly available datasets or is synthetically generated, which samples do not represent any real person. No other human subjects are involved in the study. Necessary agreements, data protection strategies and consents have already been collected. Also, our work is intended to provide a fair tool for assessing the accuracy of reconstruction methods rather than proposing a new reconstruction method.  

There are still some potential risks to mention, even though we believe they are minimal as compared to the potential advantages for the research community. A more accurate evaluation benchmark in which the real quality of a reconstruction is better assessed could lead to faster development of very accurate reconstruction methods. This, in turn, might lead to the ability of malicious users to obtain additional personal information (3D face structure) from images, without the users consent. Indeed, the 3D facial structure represents a very personal piece of information that can be used to create fake data of specific individuals. However, this risk is a common characteristic of all face reconstruction approaches. On the other hand, our method actually constitutes itself a risk mitigation technique. Our proposed tool could be used to compare two different 3D face meshes of a specific individual, so to accurately asses whether the two are similar enough to be considered to real samples, avoiding to some extent the risk of incurring into fake generated data.

{\small
\bibliographystyle{ieee}
\bibliography{main_arxiv}
}

\section*{Appendix A: Structure and Functionalities of M3D-FB}
\setcounter{figure}{0}
\setcounter{equation}{0}
\renewcommand{\theequation}{A.\arabic{equation}}
\renewcommand{\thefigure}{A.\arabic{figure}}

\begin{figure*}[h]
\begin{subfigure}[t]{0.5\textwidth}
\includegraphics[scale=1.0]{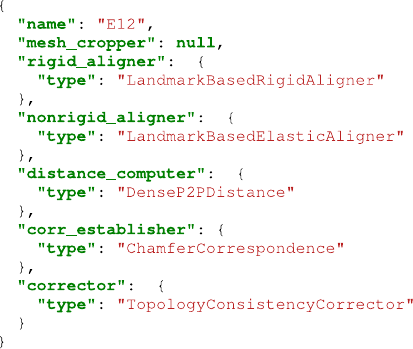}

\end{subfigure}
\begin{subfigure}[t]{0.5\textwidth}
\includegraphics[scale=1.0]{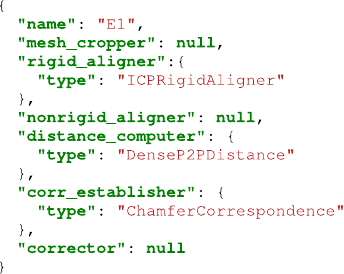}
\end{subfigure}
\caption{JSON files that implement, respectively, the error estimator E12 (left) and E1 (right).}
\label{fig:estimators}
\end{figure*}

The goal of the M3D-FB framework is to implement a 3D reconstruction error estimator by simply preparing a JSON file that specifies the chosen components~(Figure~\ref{fig:estimators},\ref{fig:rigid_alignment}). Users can also design and test alternatives for each component. Further, a complete experimental analysis with multiple reconstruction methods and/or error estimators can also be done by simply curating a JSON file~(Figure~\ref{fig:experiment_json}).  Built-in tools for producing tables and figures are provided to facilitate the inspection and evaluation of reconstruction methods or error estimators~(Figure~\ref{fig:visualization}), while reducing computational overhead by caching and re-using the computed per-vertex errors. \textit{[The code is currently provided in a .zip file together with this supplementary material, and will be released on an open-source repository should the paper be accepted for publication.]}

\paragraph{Implementing an existing or custom error estimator.} 
M3D-FB is designed in a way that all details needed to implement an estimator are added to a JSON file, and one does not have to write or modify a programming script. For example, the estimator E12 of the experiments and the standard ICP+Chamfer (E1) estimator can be implemented as in Figure~\ref{fig:estimators}.

\noindent Also, users can define options for each step. For example, rigid landmark-based registration~(RLR) is performed by default using the outer eye corners, inner eye corners and the nose tip, which correspond to the landmarks 13, 19, 28, 31 and 37 of the iBUG-51 model\footnote{\url{https://ibug.doc.ic.ac.uk/resources/300-W/}}. One can add, say, the inner mouth corners, by updating the field corresponding to rigid alignment as in Figure~\ref{fig:rigid_alignment}:
\begin{figure}[h!]
\includegraphics[scale=1.0]{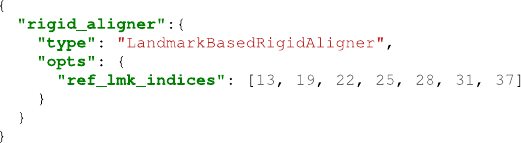}
\caption{Illustration of how a step can be modified by adding options, through an example where reference landmarks for rigid landmark alignment are specified.}
\label{fig:rigid_alignment}
\end{figure}

Each of the error computation steps (Figure~1 in the main document) are implemented with a Python class inherited from a base class. For example, RLR is implemented with a class named \texttt{LandmarkBasedRigidAligner}, which is derived from the class \texttt{BaseRigidAligner}. A user who implements his/her own rigid aligner in a class named \texttt{CustomRigidAligner} can easily test if it provides improvement over the estimators E01 or E12 by replacing the rigid aligner type in Figure~\ref{fig:estimators} with \texttt{CustomRigidAligner}. 

\paragraph{Running experiments and reporting results.}

All experimental analyses in M3D-FB are conducted with the same script, namely \texttt{run.py}, which has three arguments:
\texttt{python run.py JSON\_FILE DATA\_DIR --num\_processes NUM\_PROCESSES}
The second argument is the path that contains all necessary data, listed as explained by the end of this section. The optional \texttt{num\_processes} argument controls the number of CPU cores to be used during experiments, and setting it to values larger than 1 enables parallelization across dataset subjects. The first argument is the path to a JSON file that contains all details needed to run the experiment, including the reconstruction methods that are being compared, the error estimators that are being implemented, as well as the type of the report that will be generated. An example JSON file is shown in Figure~\ref{fig:experiment_json}.

\begin{figure}[h!]
\includegraphics[scale=1]{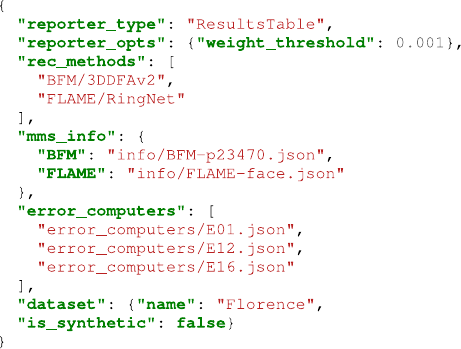}
    \caption{JSON file illustrating how an experimental analysis can be carried out.}
    \label{fig:experiment_json}
\end{figure}
The file in Figure~\ref{fig:experiment_json} produces a table (see field \texttt{reporter\_type}) with the errors of the 3DDFAv2 and RingNet methods on the Florence dataset, produced with the estimators E1, E12 and E16. Of note, once per-vertex error for a subject is estimated, it is cached for re-using later. As such, various reports can be generated without re-computing error~(Figure~\ref{fig:visualization}). For example, Figure~6 in the main text was produced by re-using the errors computed for Table~2. Moreover, the results in the bottom row of Figure~6 were also produced by re-using the same results, but reporting only on the 14,000 vertices corresponding on the inner face rather than all vertices. Users can create their own visualizations by extending the class \texttt{BaseReporter}, which enables them to minimize boilerplate code and leverage the error caching functionality.

\begin{figure*}
    \centering
    \includegraphics[width=0.98\linewidth]{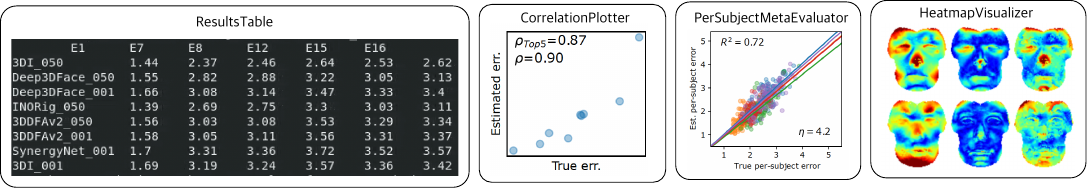}
    \caption{Examples to currently supported result reports and visualizations. }
    \label{fig:visualization}
\end{figure*}

The reconstruction methods in Figure~\ref{fig:experiment_json} are preceded with an abbreviation corresponding to the mesh topology that the method is based on (\eg, BFM or FLAME); and the paths to the JSON in the field \texttt{mms\_info} contain the necessary information related to the mesh topology. As such, the error of a reconstruction method can be estimated in a standardized way across reconstruction methods and without the need to modify any python files or scripts. For example, the JSON file named \texttt{BFM-p23470.json} contains the landmark indices \wrt the mesh cropping of 23,470 points (Section~4.1), which are needed for rigid registration or non-rigid warping. The files \texttt{BFM-p23470.json} and \texttt{FLAME-face.json} are included in the current version of the codebase for convenience. 

All reconstructed meshes and ground truth scans for a dataset are expected to be in the same directory (\ie, \texttt{DATA\_DIR}). It is assumed that the ground truth meshes and landmarks of, say, the first subject of the Florence dataset, are located below

\texttt{DATA\_DIR/Florence/Gmeshes/id0001.txt} 

\texttt{DATA\_DIR/Florence/Gmeshes/id0001.lmks}

The reconstructed meshes are listed in a nested directory, structured to specify the mesh topology (\eg, BFM or FLAME) used by the reconstruction method; and the cropping that is used (\eg, the 23,470-point version of BFM) during experiments. For example, the reconstructed mesh for the first subject of the Florence dataset obtained with the 3DDFAv2 method is to be listed at:
\texttt{/Rmeshes/BFM/p23470/3DDFAv2/id0001.txt}
A Python script is provided to allow users crop reconstructed meshes according to this version of BFM with the 23,470 points, or the face vertices of the FLAME model.

\section*{Appendix B: Elastic Landmark Registration}
\setcounter{figure}{0}
\setcounter{equation}{0}
\renewcommand{\theequation}{B.\arabic{equation}}
\renewcommand{\thefigure}{B.\arabic{figure}}

Elastic landmark registration (ELR) is a non-rigid mesh warping approach that uses no information other than landmarks on the reconstructed mesh and the corresponding ground truth scan. This process is also referred to as landmark-based pre-alignment~[30], but we use the term ELR to distinguish it from the rigid landmark-based alignment~(Section~3). For completeness, we summarize this procedure below.

Suppose that $g'_{l_1},g'_{l_2},\dots,g'_{l_L}$ are $L$ annotated 3D landmark points on the ground truth scan, and that $r_{l_1}, r_{l_2}, \dots, r_{l_L}$ are the $L$ corresponding points on the reconstructed mesh $R$. The goal of ELR is to warp the entire reconstructed mesh $R$ in a way that the discrepancy between the corresponding landmarks is minimized. This is achieved by computing a per-landmark movement vector $u_{i}$. ELR assumes that, whenever a landmark point $r_{l_i}$ is moved by $u_{i}$ to the point $r_{l_i}+u_{i}$, another point $r_k$ on the mesh $R$ will also move to:
\begin{equation}
r_k+\alpha_{ki}u_{i},
\label{eq:drag}
\end{equation}
\noindent
where $\alpha_{ki}$ is the corresponding \textit{magnitude coefficient}. The coefficient $\alpha_{ki}$ is determined based on the distance between $r_k$ and $r_{l_i}$. That is, the closer the point $r_k$ is to  $r_{l_i}$ the more it moves. Furthermore, it is assumed that the mesh point that is farthest from $r_{l_i}$ will not be affected by this movement at all. Thus, the $\alpha_{ki}$ coefficient is set as
\begin{equation}
\alpha_{ki} = 1-\frac{||r_k-r_{l_i}||}{\max_j ||r_j -r_{l_i}||},
\label{eq:mag}
\end{equation}
where $||\cdot||$ is the Euclidean norm. 

According to~\eqref{eq:drag}, the movement of any landmark $r_{l_i}$ affects also other landmark points $r_{l_j}$ on the mesh. Therefore, one should determine the vectors $u_1, u_2, \dots, u_L$ simultaneously so that \textit{all} landmark points $r_{l_1}, r_{l_2}, \dots, r_{l_L}$ on the reconstructed mesh are aligned to their counterparts $g'_{l_1}, g'_{l_2}, \dots, g'_{l_L}$. Let us define $A$ as an $N \times L$ matrix whose $ij$-th element is $\alpha_{ij}$, and $\tilde{A}$ as the $L \times L$ submatrix corresponding to the rows of $A$ with indices $l_1, l_2, \dots, l_L$. Then, the per-landmark movement vectors needed for ELR are obtained by solving the linear system
\begin{equation}
\tilde{A} U = E,\label{eq:reg_params}
\end{equation}
for $U$, where $U=(u_1^T, u_2^T, \dots, u_L^T)^T$ is the $L\times 3$ matrix whose rows are the movement vectors that we seek, and $E$ is the $L\times 3$ matrix whose $i$th row is $g'_{l_i}- r_{l_i}$. Finally, the reconstructed mesh $R'$ that is non-rigidly aligned to the ground truth can be obtained by
\begin{equation}
R'=R+AU. \label{eq:reg}
\end{equation}
\noindent

\section*{Appendix C: Duplicate points}
\setcounter{figure}{0}
\setcounter{equation}{0}
\renewcommand{\theequation}{C.\arabic{equation}}
\renewcommand{\thefigure}{C.\arabic{figure}}

The design of the correction step (Section~3 paragraph 6 of the main text) is motivated by the observation that fractures arise in the matched ground truth $\hat{G}$ after establishing point correspondences using the Chamfer, \ie, nearest-neighbor criterion. Such fractures are due to the fact that two or more distinct points on $R$ (or $G$) are assigned to the same point in the other point cloud, causing irregularities in the matched point cloud. There is indeed no constraint that prevents it to happen.

\begin{figure}[!t]
    \centering
    \includegraphics[scale=0.183]{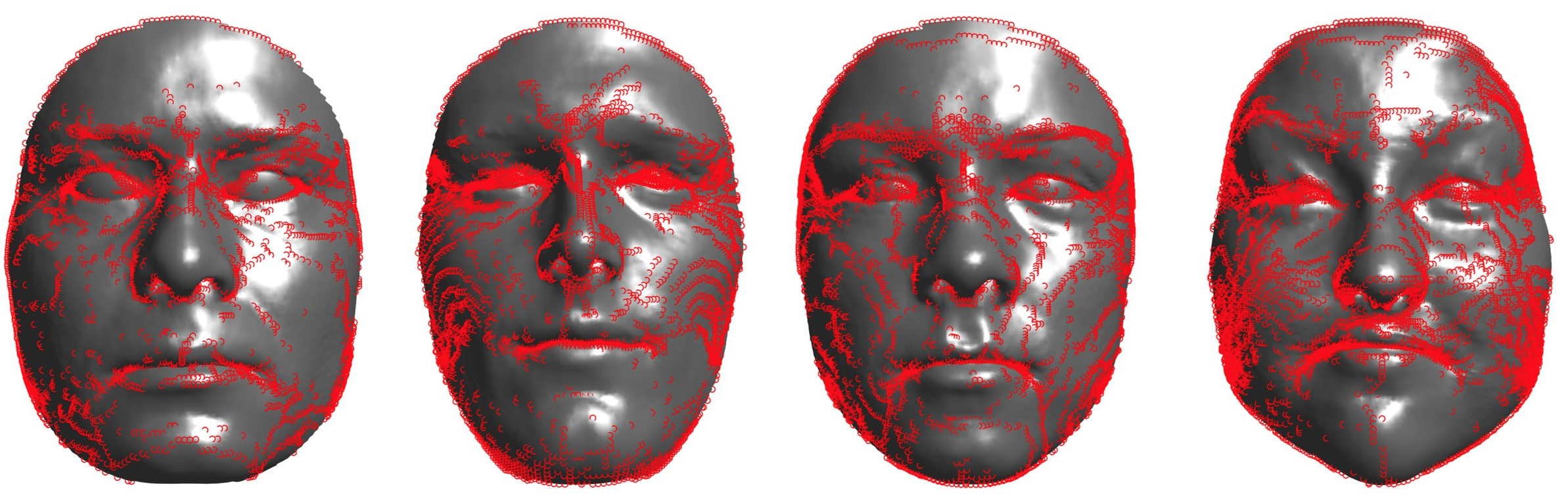}
    \caption{Examples of duplicate points association. Red points in the displayed ground truth meshes $G$ represent indices of points that are associated to multiple points in $R$, leaving a certain number of points in $G$ unmatched ultimately causing uneven point distributions.}
    \label{fig:duplicates}
\end{figure}

In Table~\ref{tab:duplicates}, to highlight the severity of this problem and how it leads to biased error measures, we report the average number of non-unique point associations as resulting from using the Chamfer criterion. It turns out that for all the methods, an average of more than the 10\% of points are matched to a point for which a nearest neighbor association has already been established. More precisely, the average number of point duplicates computed on the 100 subjects of the \textit{BFM} dataset is of 3213, 3243, 2945, 3586, and 3320 for 3DI, 3DDFAv2, Deep3DFace, INORig and the average BFM face, respectively (the BFM meshes have 23,470 points in total). Figure~\ref{fig:duplicates} displays some examples. The reader can easily appreciate that this issue is rather severe, mostly because the regions where an incorrect point-to-point matching is established are in correspondence of relevant face areas such as the eyes, or the mouth.

\begin{table}[!h]
    \centering\small
    \caption{Average percentage of duplicate nearest neighbors.}
    \begin{tabular}{c|c|c|c|c} \hline
        3DI	&	3DDFAv2	&	Deep3DFace	&	INORig	&	meanface	\\ \hline
        13.7\%	&	13.8\%	&	12.5\%	&	15.2\%	&	14.1\%	\\ \hline
    \end{tabular}
    \label{tab:duplicates}
\end{table}

\end{document}